\documentclass[letterpaper, 10 pt, conference]{ieeeconf}  
\IEEEoverridecommandlockouts   

\usepackage{subfig,float}

\usepackage[backend=biber,sorting=none]{biblatex}
\usepackage{amsmath,amssymb,amsfonts}
\usepackage{algorithmic}
\usepackage{graphicx}
\usepackage{booktabs}

\usepackage{textcomp}
\usepackage{xcolor}
\def\BibTeX{{\rm B\kern-.05em{\sc i\kern-.025em b}\kern-.08em
    T\kern-.1667em\lower.7ex\hbox{E}\kern-.125emX}}

\usepackage{upgreek}
\usepackage{mathtools}
\usepackage{stmaryrd}
\usepackage{algorithm}
\usepackage{comment}

\newtheorem{remark}{Remark}

\begin{document}
%
\title{\LARGE \bf Safe Obstacle-Free Guidance of Space Manipulators in Debris Removal Missions via Deep Reinforcement Learning}
%
%
%

\author{Vincent Lam$^{1}$ and Robin Chhabra$^{1}$
\thanks{*This work was supported by a grant from the Natural Sciences and Engineering Research Council of Canada (DGECR-2019-00085).}
\thanks{$^{1}$Embodied Learning and Intelligence for eXploration and Innovative soft Robotics (ELIXIR) Lab, Toronto Metropolitan University, Toronto, ON, Canada.
        {\tt\small 	hienvinh.lam@torontomu.ca}, {\tt\small robin.chhabra@torontomu.ca}}%
}

\maketitle

\begin{abstract}
The objective of this study is to develop a model-free workspace trajectory planner for space manipulators using a Twin Delayed Deep Deterministic Policy Gradient (TD3) agent to enable safe and reliable debris capture. A local control strategy with singularity avoidance and manipulability enhancement is employed to ensure stable execution. The manipulator must simultaneously track a capture point on a non-cooperative target, avoid self-collisions, and prevent unintended contact with the target.  
To address these challenges, we propose a curriculum-based multi-critic network where one critic emphasizes accurate tracking and the other enforces collision avoidance. A prioritized experience replay buffer is also used to accelerate convergence and improve policy robustness. The framework is evaluated on a simulated seven-degree-of-freedom KUKA LBR iiwa mounted on a free-floating base in Matlab/Simulink, demonstrating safe and adaptive trajectory generation for debris removal missions.
\end{abstract}


\IEEEpeerreviewmaketitle

\section{Introduction}

Decades of human space activity have transformed low Earth orbit into a heavily congested environment, with nearly 6000 tons of debris from defunct satellites, rocket bodies, and spacecraft fragments~\cite{NASA,NASA2,Capstone}. This growing debris population poses a significant risk to the safety and sustainability of future missions. Active debris removal has thus emerged as a critical challenge, with space manipulators offering one of the most promising approaches for capturing and deorbiting non-cooperative targets. In such missions, a free-floating manipulator must not only track a capture point on the target with high precision but also maintain a safe distance to enable a reliable capture phase.

Trajectory planning for robotic manipulators has been studied extensively in terrestrial settings. Classical approaches include virtual potential fields~\cite{Khatib}, configuration-space planning~\cite{C-space}, and graph- or swarm-based methods such as A* search~\cite{A_starred} and swarm intelligence~\cite{Swarm}. While effective on fixed-base ground manipulators, these methods do not translate well to free-floating space systems. The main difficulty arises from the manipulator's uncertain interaction and its strong dynamic coupling with the spacecraft base, which significantly complicates both planning and control~\cite{Yoshida,Papadopoulos,Capstone2}. 
Given these challenges, Deep Reinforcement Learning (DRL) has emerged as a compelling alternative. DRL enables agents to learn policies directly through interaction, without requiring an explicit model of the environment. Recent advances have demonstrated its potential in robotic manipulation. For instance, Soft Actor-Critic (SAC) has been applied to redundant manipulators for obstacle avoidance~\cite{redundant_obstacle_avoid}, while the Deep Deterministic Policy Gradient (DDPG) algorithm~\cite{DDPG} has been widely adopted for trajectory planning of space manipulators in joint space~\cite{york_jointspace,virtual_field_jointspace,DDPG2}. However, DDPG suffers from overestimation bias, which often leads to unstable learning. To address this issue, the Twin Delayed DDPG (TD3) algorithm introduces dual critics and delayed policy updates, yielding faster convergence and more reliable performance~\cite{TD3_paper}.  

Although RL has achieved success in solving individual tasks, space debris removal requires managing multiple objectives simultaneously, such as accurate tracking of a capture point and collision avoidance. Multi-critic architectures provide a natural extension to address such requirements. Instead of relying on a single critic to evaluate all objectives, separate critics can be assigned to distinct tasks. This idea has been successfully applied across diverse domains, including training agents with multiple combat strategies in video games~\cite{UFC}, path planning for autonomous underwater vehicles~\cite{Submarine}, and decentralized navigation of multiple robots in congested traffic-like environments~\cite{Multi_Critic_TD3}.  
Another important development in RL is the use of prioritized experience replay (PER). Unlike standard replay buffers, PER assigns higher sampling probability to transitions with greater learning value. This accelerates convergence by focusing updates on informative experiences. DeepMind demonstrated the effectiveness of PER in combination with Deep Q-Networks (DQN) and Double DQN, achieving state-of-the-art performance on Atari benchmarks~\cite{GoogleDeepMind}.  

In this work, we build on our previous research~\cite{Timothy,Timothy2} to develop a safe trajectory planner for a seven-degree-of-freedom free-floating space manipulator in a three-dimensional simulation environment. Our approach leverages a TD3 agent with a dual-critic architecture, where one critic specializes in tracking while the other focuses on collision avoidance. To further improve learning efficiency and robustness, we incorporate a novel prioritized experience replay into the training framework.  
The contributions of this paper are summarized as follows:
\begin{itemize}
    \item We propose a TD3 agent for obstacle-free workspace trajectory planning of a space manipulator on the Special Euclidean group $\mathbb{SE}(3)$, leveraging safe control frameworks previously developed in~\cite{Borna_Control2,Borna_Control,Rousso}.
    \item We introduce a multi-critic TD3 architecture in which tracking and collision avoidance are learned sequentially by separate critics, improving task specialization and policy performance.
\end{itemize}

The remainder of the paper is organized as follows. Section~2 formulates the debris removal problem and describes the control framework. Section~3 introduces the TD3 algorithm. Section~4 presents the proposed trajectory planner and training methodology. Section~5 reports the simulation results. Section~6 concludes the paper and highlights key findings.

\section{Problem Statement}
\subsection{Preliminary}

The free floating space manipulator in 3D space is modeled as a base spacecraft which has an $n$-link manipulator mounted. All joints of the manipulator are assumed to be revolute. To be able to generate velocity in any arbitrary direction in 3D space, we assume that $n \ge 6$. The properties of the manipulator (link dimensions, mass, inertia, etc) are collected into a vector $\rho$. In a multibody system, the pose of frame $i$ relative to frame $j$ is defined by $g^j_i = 
\begin{bmatrix}R^j_i & P^j_i \\ \mathbb{O}_{1\times3} & 1 \end{bmatrix} \in \mathbb{SE}(3)$,
where $R^j_i \in \mathbb{SO}(3)$ and $P^j_i \in \mathbb{R}^3$ define the relative orientation and position of frame $i$ with respect to frame $j$. $\mathbb{O}$ denotes the zero matrix with the specified dimensions.
    
Let $I,0$ denote the inertial frame and the frame attached to the base of the manipulator. Hence $g^I_0 \in \mathbb{SE}(3)$ denotes the pose of the manipulator base with respect to the inertial frame. Let $\phi \in \mathbb{R}^n$ denote the joint angles of the manipulator and $\xi_i \in \mathbb{R}^6$ denote the screw axis of the joint $i$ expressed with respect to the base frame $0$ in the zero configuration of the manipulator ($\phi = \mathbb{O}_{n \times 1}$). The relative pose at the joints $i \in (1,...,n+1)$ of the manipulator with respect to the manipulator base frame $0$, is given by the forward kinematics mapping $F_i(\phi|\rho) \in \mathbb{SE}(3)$, note that $i = n + 1$ corresponds to the end effector frame. $F_i(\phi|\rho)$ is defined by the following 

\begin{equation}
    g^0_i = e^{{[\xi_1]}\phi_1}...e^{{[\xi_i]}\phi_i}\overline{g}^0_i,
\end{equation}
where $[{\bullet}]$ denotes the $se(3)$ matrix representation of a given velocity vector $\bullet$, $\overline{g}^0_i$ denotes the pose of the joint frame $i$ with respect to the base frame in the zero configuration of the manipulator. The pose of the joint $i$ can be transformed into $I$ by premultiplying by $g^I_0$

\begin{equation}
    g^I_i = g^I_0 g^0_i.
\end{equation}

Using (2), the pose of the end effector $g^I_{n + 1} = g^I_{ee}$ can be computed. The mass matrix of the space manipulator system is given by 

\begin{equation}
    M(\phi|\rho) = 
    \begin{bmatrix}
        M_b & M_{bm}\\ M^T_{bm} & M_m
    \end{bmatrix}
\end{equation}
where the submatrices of $M$ are $M_b \in \mathbb{R}^{6 \times 6}$ which denotes the mass matrix of the spacecraft, $M_m \in \mathbb{R}^{n \times n}$ denotes the mass matrix of the manipulator, and $M_{bm} \in \mathbb{R}^{6 \times n}$ denotes the coupling mass matrix between the spacecraft and the manipulator. The space manipulator is assumed to be free floating, hence the conservation of momentum can be assumed. The total momentum of the space manipulator system, assuming zero initial momentum, is given by 

\begin{equation}
    M_bV^I_b + M_{bm}\dot{\phi} = \mathbb{O}_{6 \times 1},
\end{equation}
where $V^I_b \in \mathbb{R}^6$ denotes the velocity of the base spacecraft in $I$. Rearranging (4) to solve for $V^I_b$ gives

\begin{equation}
    V^I_b = J_{bm}\dot{\phi},
\end{equation}
where $J_{bm} \in \mathbb{R}^{6 \times n}$ is the coupling Jacobian matrix between the base spacecraft and the manipulator. Let $v^I_{ee}, \omega^I_{ee} \in \mathbb{R}^3$ denote the linear and angular velocity of the end effector in $I$. The end effector velocity is the result of the contribution of the joint velocities and motion of the base spacecraft

\begin{equation}
    \begin{bmatrix}
    v^I_{ee} \\
    \omega^I_{ee}
    \end{bmatrix} = J_b
    V^I_b + J_m \dot{\phi},
\end{equation}
where $J_b \in \mathbb{R}^{6 \times 6}$ denotes the base spacecraft Jacobian and $J_m \in \mathbb{R}^{6 \times n}$ denotes the manipulator Jacobian. Substituting (5) into (6) gives

\begin{equation}
    \begin{bmatrix}
    v^I_{ee} \\
    \omega^I_{ee}
    \end{bmatrix} = (J_b J_{bm} + J_m)\dot{\phi} = J_{ee} \dot{\phi},
\end{equation}
such that
$J_{ee} \in \mathbb{R}^{6 \times n}$ is known as the generalized Jacobian of the space manipulator, which maps the joint velocities of the manipulator directly to the velocity of the end effector while taking into consideration the spacecraft's motion.

The capture point is modeled as a sphere with a radius $\zeta$ that is rigidly connected to a rectangular base and moves randomly through space. We consider $g^I_{tar},g^I_{o} \in \mathbb{SE}(3)$ to denote the pose of the center of the sphere and the pose of the center of the rigidly connected base, respectively.  

The following sections provide an overview of the controller developed for the space manipulator. Further details can be found in \cite{Borna_Control2} \cite{Borna_Control}.

\subsection{Local Manipulator Controller}

\subsubsection{Workspace PID Control}
In this section, an extended PID controller developed on Lie groups is proposed to control the end effector to track a generated trajectory by TD3. Let frame $t$ be the target pose for the end effector. Consider the pose of the end effector expressed with respect to the target frame $g^{t}_{ee} \in \mathbb{SE}(3)$ be the output of the manipulator system, where $g^{t}_{ee} = 
\begin{bmatrix}R^t_{ee} & P^t_{ee} \\ \mathbb{O}_{1\times3} & 1 \end{bmatrix} \in \mathbb{SE}(3)$. The relative pose error function $E_g$ is given by

\begin{equation}
    E_g = \frac{1}{2} \boldsymbol{tr}(\mathbb{I}_{3 \times 3} - R^t_{ee}) + \frac{1}{2}(P^t_{ee})^TP^t_{ee},
\end{equation}
where $\boldsymbol{tr}(\bullet)$ represents the trace of the matrix $\bullet$. Note that $E_g$ is a coordinate-free positive definite and symmetric pose error function on $\mathbb{SE}(3)$. $E_g$ equals zero if and only if $g^t_{ee} = \mathbb{I}_{4 \times 4}$. $E_g$ behaves locally quadratically when $E_g$ is below some critical threshold $E_{lim} > 0$. The gradient of the relative pose error $\nabla E_g$ is given by taking the time derivative of $E_g$ and using properties of the trace function

\begin{equation}
    \dot{E_g} = (\nabla E_g)^T \prescript{B}{}{V^t_{ee}} = \begin{bmatrix}
        P^t_{ee} \\ \boldsymbol{sk}(R^t_{ee})^\vee
    \end{bmatrix}^T \prescript{B}{}{V^t_{ee}},
\end{equation}
where $\prescript{B}{}{V^t_{ee}} \in \mathbb{R}^6$ denotes the body velocity  of the end effector with respect to the target frame $t$, the notation $\prescript{B}{}{V^j_i}$ denotes the relative velocity of frame $i$ with respect to frame $j$ but expressed in frame $i$. The PID control law used to stabilize the output of the system $g^t_{ee} \xrightarrow{} \mathbb{I}_{4 \times 4}$ is given by the following

\begin{equation}
    U_{pid} = -K_p \nabla E_g (g^t_{ee}) - K_d \prescript{B}{}{V^t_{ee}} - K_iF_i \in \mathbb{R}^6
\end{equation}

\begin{equation}
    \dot{F_i} = K_p \nabla E_g (g^t_{ee}) + K_d \prescript{B}{}{V^t_{ee}}
\end{equation}
where $K_p,K_d,K_i \in \mathbb{R}^{6 \times 6}$ denote the matrices that represents the proportional, derivative, and integral gains. 

The PID controller $U_{pid}$ provides local asymptotic stability of the system output $g^t_{ee}$ near $\mathbb{I}_{4 \times 4}$ if the following conditions are satisfied, $a^{'}_2 a^{'}_3 > (a_3)^2$ and $a^{'}_2 (a_3)^2 > 4a^{'}_1 a^{'}_3 + (a_2)^2 a^{'}_3$ where $a_1, a_2, a_3$ and $a^{'}_1, a^{'}_2, a^{'}_3$ denote the largest and smallest eigenvalues of the matrices $K_p,K_d,K_i$.
Finally, a feedback transformation is implemented to linearize output, further details of the derived feedback transformation can be found in \cite{Borna_Control2} \cite{Borna_Control}.

\subsubsection{Nullspace Control}

This section introduces the nullspace component of the controller, consider the following 

\begin{equation}
    \boldsymbol{\lambda}(\phi) = \sqrt{\det(J_{ee}J_{ee}^T)}
\end{equation}
where $\boldsymbol{\lambda}(\phi) \in \mathbb{R}$ is the manipulability parameter that quantifies how close the manipulator is to singularity. We assume that the manipulator at hand has at least 7 independent joints, hence we can use internal motion to keep the manipulator dexterous without compromising tracking in taskspace. The partial derivative of $\boldsymbol{\lambda}(\phi)$ is given by the following

\begin{equation}
    \nabla \boldsymbol{\lambda} = 
    \begin{bmatrix}
        \frac{\partial\boldsymbol{\lambda}}{\partial\phi_1}...\frac{\partial\boldsymbol{\lambda}}{\partial\phi_n}
    \end{bmatrix}^T,
\end{equation}

\begin{equation}
    \frac{\partial \boldsymbol{\lambda}}{\partial \phi_i} = \frac{\boldsymbol{\lambda}}{2}
    \boldsymbol{tr}((J_{ee}J^T_{ee})^{-1}(J_{ee} \frac{\partial J^T_{ee}}{\partial \phi_i} + \frac{\partial J_{ee}}{\partial \phi_i}J^T_{ee})).
\end{equation}

Let $J^{\dagger}_{ee} \in \mathbb{R}^{n \times 6}$ denote the Moore Penrose Pseudo Inverse. The nullspace projection matrix $\mathbb{N}$ is given by the following

\begin{equation}
    \mathbb{N} = \mathbb{I}_{n \times n} - J^{\dagger}_{ee} J_{ee} \in \mathbb{R}^{n \times n}
\end{equation}
where $\mathbb{I}$ denotes identity matrix with specified dimensions. Introducing the matrix $\mathbb{Z} \in \mathbb{R}^{n \times (n-6)}$ whose column vectors represent the linearly independent basis vectors that span the nullspace of $J_{ee}$. Note that since the nullspace varies depending on the configuration of the manipulator $\phi$, hence $\mathbb{Z}$ is only defined locally. The first nullspace control law is given by the following

\begin{equation}
    U_{ns_1} = K_{\boldsymbol{\lambda}}\mathbb{Z}^T\nabla \boldsymbol{\lambda} \in \mathbb{R}^{n - 6}
\end{equation}
where $K_{\boldsymbol{\lambda}} \in \mathbb{R}^{(n - 6) \times (n-6)}$ is a positive symmetric matrix. $U_{ns_1}$ provides the manipulator with a push away from singularity configurations. Since $\mathbb{Z}$ contains the basis vectors that define the nullspace, let $v \in \mathbb{R}^{n - 6}$ such that $\mathbb{N} \dot{\phi} = \mathbb{Z}v$. Hence, $v$ represents the magnitude of the internal motion of the manipulator. The second nullspace control is 

\begin{equation}
    U_{ns_2} = -K_v v \in \mathbb{R}^{n-6}
\end{equation}
where $K_v \in \mathbb{R}^{(n-6) \times (n -6)}$ is a positive symmetric matrix. $U_{ns_2}$ acts as a spring that pulls the vector $v$ towards $\mathbb{O}_{(n-6) \times 1}$, thus stabilizing the internal motion of the space manipulator. Combining (16) and (17) gives the complete nullspace control law that pushes the space manipulator towards dexterous configurations while keeping the internal motion stable and bounded. 

\begin{equation}
    U_{ns} = U_{ns_1} + U_{ns_2}
\end{equation}

\subsubsection{Singularity Avoidance}

The nullspace control law $U_{ns}$ provides the space manipulator resistance near the vicinity of singularity configurations, but when the manipulator is operating near the region of its workspace limits, $U_{ns}$ is insufficient in avoiding singularity. An additional singularity avoidance controller utilizing virtual potential field is proposed. The potential field $\mathbb{U}_{\boldsymbol{\lambda}}$ is defined as follows

\begin{equation} 
       \mathbb{U}_{\boldsymbol{\lambda}} = \begin{cases} 
            \frac{1}{2}(1 - \frac{\boldsymbol{\lambda}_{lim}}{\boldsymbol{\lambda}})^2, & \text{if } \boldsymbol{\lambda} \le \boldsymbol{\lambda}_{lim}  \\
            0 , & \text{if } \boldsymbol{\lambda} > \boldsymbol{\lambda}_{lim}
        \end{cases}
\end{equation}where $\boldsymbol{\lambda}_{lim} \in \mathbb{R}$ is a threshold that determines when the singularity avoidance is activated. Note that this virtual field will explode when $\boldsymbol{\lambda}$ = 0. The Lipschitz continuous gradient of $\mathbb{U}_{\boldsymbol{\lambda}}$ with respect to $\phi$ is given by the following

\begin{equation} 
       \nabla \mathbb{U}_{\boldsymbol{\lambda}} = \begin{cases} 
            \frac{\boldsymbol{\lambda}_{lim}}{\boldsymbol{\lambda}^2}(1 - \frac{\boldsymbol{\lambda}_{lim}}{\boldsymbol{\lambda}}) \nabla \boldsymbol{\lambda}, & \text{if } \boldsymbol{\lambda} \le \boldsymbol{\lambda}_{lim}  \\
            \mathbb{O}_{n \times 1} , & \text{if } \boldsymbol{\lambda} > \boldsymbol{\lambda}_{lim}
        \end{cases}
\end{equation}
The singularity avoidance controller expressed in the taskspace of space manipulator is given by

\begin{equation}
    U_s = - (J^{\dagger}_{ee})^T \nabla \mathbb{U}_{\boldsymbol{\lambda}} \in \mathbb{R}^6.
\end{equation}

The final taskspace controller is the summation between the PID controller $U_{pid}$ and the singularity avoidance controller $U_s$

\begin{equation}
    U_{ts} = U_{pid} + U_s \in \mathbb{R}^6
\end{equation}

As long as the target pose is within the dexterous space of the space manipulator, $U_s$ will not be activated and the end effector will track the desired trajectory using $U_{pid}$. If the target pose enters the workspace limits of the space manipulator or even traverses beyond reach, then $U_s$ will be activated and prevent the manipulator from regions of lower manipulability by halting the motion of the space manipulator in the direction of $\nabla \boldsymbol{\lambda}$. 

\section{The TD3 Agent}

The TD3 agent is a state-of-the-art DRL algorithm. Let $s \in S$ denote the current state of the environment, and let $a \in A$ denote the action the TD3 agent performs given the current state $s$. The architecture of the TD3 agent is composed of a deterministic actor network $\mu$(s), two state-action Q value critic networks $Q_1(s,a)$, $Q_2(s,a)$ where each network is parameterized by the weights $\theta$, $\phi_1$, $\phi_2$ respectively. Each network has a target network counterpart that has identical network structures $\mu'$(s), $Q'_1(s,a)$, $Q'_2(s,a)$ each parameterized by the weights $\theta'$, $\phi'_1$, $\phi'_2$, respectively. 

At a given timestep t, the agent at state $s(t)\in S$ outputs the set of deterministic action $a(t)\in A$ based on the policy function which is approximated by the actor network $\mu(s)$. Noise $\epsilon_1$ is added to the action $a(t)$ for exploration purposes. The action $a(t)$ is subsequently applied to the environment to obtain the reward $r(t)$ and next state $s(t+1)$. The experience tuple is collected as $(s(t),a(t),r(t),s(t+1))$ and stored in a replay buffer $\mathcal{B}$. During the learning process, a mini-batch of size N experience tuples is formed by sampling from $\mathcal{B}$. The weights of each critic network $\phi_1$, $\phi_2$ are updated by gradient descent,  minimizing the following loss function for $k = 1,2$

\begin{equation}
    L_k = \frac{1}{N}\sum_{i=1}^{N}({y_i - Q_k(s_i,a_i|\phi_k)})^2
\end{equation}
where the optimal target Q value $y_i$ is equal to the reward $r_i(t)$ for terminal next states, for non-terminal next states, the optimal target Q value is given by the Bellman Ford equation

\begin{equation}
    y_i = r_i(t) + \gamma \min_k(Q'_k(s(t+1),clip(\mu'(s(t+1)|\theta') + \epsilon_2)|\phi'_k))
\end{equation}
where $\gamma$ is the reward discount factor, $\epsilon_2$ denotes the noise added to the action produced by the target actor $\mu'(s)$ for the purpose of target policy smoothing. $clip(\bullet)$ denotes clipping all the elements of a vector $\bullet$ between a upper and lower limit, where in this case the limits are given by the range of values of the action space $A$. $min_k$ denotes taking the minimum Q value outputted by both target critic $Q'_k(s,a)$. The actor network weights $\theta$ are updated by gradient ascent, maximizing the given function

\begin{equation}
    \nabla_\theta J = \frac{1}{N}\sum_{i=1}^N\nabla_a \min_k(Q_k(s,a|\phi_k)|_{s=s_i,a=\mu(s_i)})\nabla_\theta \mu(s|\theta)|_{s=s_i}
\end{equation}

\begin{equation}
    J = \frac{1}{N}\sum_{i=1}^N\min_k(Q_k(s,a|\phi_k)|_{s=s_i,a=\mu(s_i)})
\end{equation}

The weights of the corresponding target networks $\theta'$, $\phi'_1$, $\phi'_2$' are updated periodically to improve learning stability using the following equations, where $\tau <= 1$ is known as the smoothing factor.

\begin{equation}
    \phi'_k = \tau\phi_k + (1 - \tau)\phi'_k
\end{equation}
\begin{equation}
    \theta' = \tau\theta + (1 - \tau)\theta'
\end{equation} 

\section{Proposed Trajectory Planner}

We propose a TD3 agent to act as a trajectory planner in the space manipulator taskspace, i.e., $\mathbb{SE}(3)$. Based on the current state $s(t)$, the agent will output the end effector velocities expressed in the end effector body frame $\prescript{B}{}{V^I_{ee}} \in \mathbb{R}^6$ that are most appropriate to track the capture point while avoiding obstacles. Taking the end effector current pose $g^I_{ee}$ and velocity $\prescript{B}{}{V^I_{ee}}$, the target end effector pose $\bar g^I_{ee}$ is given by

\begin{equation}
     \bar g^I_{ee}(t+\Delta t) = g^I_{ee}(t)e^{[\prescript{B}{}{V^I_{ee}]}\Delta t}.
\end{equation}
The quantities $\bar g^I_{ee}$ and $\prescript{B}{}{V^I_{ee}}$ are then provided as desired inputs to the task-space controller in~(22).

\subsection{States and Actions}
Due to multitask requirements for safe debris removal, we divide the complete state into two substates $s_c,s_o$. The capture state $s_c$ contains information that is relevant for the end effector to track and acquire the capture point. $s_c$ contains the joint states of the manipulator, and the configuration and motion of the capture point and end effector frames. Note that all parameters about the end effector and the capture point are expressed with respect to the inertial frame $I$, for visibility the superscript $\bullet ^I$ will be dropped in this section. Let $P_{ee},v_{ee},\theta_{ee},\omega_{ee} \in \mathbb{R}^3$ denote the position vector, linear velocity, Euler angles, and angular velocity of the end effector in $I$. Let $P_{tar},v_{tar},\theta_{tar},\omega_{tar} \in \mathbb{R}^3$ denote the same parameters but for the capture point. $s_c$ is given by

\begin{equation}
    s_c = [\phi^T,\dot{\phi}^T,P_{ee}^T,v_{ee}^T,\theta_{ee}^T,\omega_{ee}^T,
P_{tar}^T,v_{tar}^T,\theta_{tar}^T,\omega_{tar}^T] =: S_c
\end{equation}

The obstacle state $s_o$ describes the geometry of the manipulator and the rectangular base that is rigidly attached to the capture point. The geometrical representation is crucial to prevent collision between the two. The geometry of the manipulator is represented by $n_L$ evenly distributed points along each link, i.e., $L_{ij}\in \mathbb{R}^3 (i = 1,...,n$ and $j = 1,...,n_L)$. The geometry of the rectangular base is represented by $n_o$ points on the surface of the base, i.e., $T_k\in \mathbb{R}^3 (k = 1,...,n_o)$. $s_o$ contains the following

\begin{equation}
    s_o = [P_o^T,L_{11}^T,...,L_{nn_L}^T,T_1^T,...,T_{n_o}^T] =: S_o
\end{equation}
where $P_o\in \mathbb{R}^3$ denotes the position of the center of the rectangular base. Combining $s_c,s_o$ gives the full state $s = [s_c^T,s_o^T]^T\in S$. The action of the agent is end effector body frame velocity, i.e. $a = {\prescript{B}{}{V^I_{ee}}}\in \mathbb{R}^6$. The $n_o$ points on the surface of the rectangular obstacle is given by

\begin{equation}
    \begin{bmatrix}
        {T}_k \\ 1
    \end{bmatrix} = {g}_o
    \begin{bmatrix}
        b_k \\ 1
    \end{bmatrix} \quad k = 1,...n_o
\end{equation}
where $b_k \in \mathbb{R}^3$ is the position vector of the $k$ surface point with respect to the center of the rectangular base. Using forward kinematics and (2), the pose of the joints and the end effector ${g}_{i}$ are obtained. From ${g}_i$ we extract the position vector of the pose ${P}_i \in \mathbb{R}^3$. The positions of the $n_L$ points along the $n$ links are calculated by

\begin{equation}
    {L}_{ij} = {P}_i + \frac{j - 1}{n_L}({P}_{i+1} - {P}_i) \quad j = 1,...,n_L
\end{equation}

\subsection{Reward Function}
The reward function is divided into two subfunctions, one corresponds to tracking $r_c$ and the other obstacle avoidance $r_o$, the total reward function is the sum of the two i.e., $r = r_c + r_o$. 

\begin{remark}
    $||\bullet||$ denotes the Euclidean norm.
\end{remark}

The capture reward function $r_c$ is composed of three components, i.e., $r_c = r_{c}^1 + r_{c}^2 + r_{c}^3$

    i) A dense punishment based on the linear distance between the end effector and the capture point:

    \begin{equation}
        r_c^1 = -k_{c1}||P_{ee} - P_{tar}||
    \end{equation}
where $k_{c1}\in \mathbb{R}^+$ is a constant scaling factor. $r_c^1$ encourages the agent to generate trajectories such that the end effector translates linearly towards the capture point to reduce the distance between the two.

    ii) A sparse reward activated when the end effector successfully makes contact with the capture point that is a sphere with radius $\zeta$:

    \begin{equation} 
       r_c^2 = \begin{cases} 
            k_{c2}, & \text{if } ||P_{ee} - P_{tar}|| \le \zeta  \\
            0 , & \text{otherwise}
        \end{cases}
    \end{equation}
where $k_{c2} \in \mathbb{R}^+$ is a constant scaling factor.  

    iii) A sparse punishment when the manipulator gets too close to singularity configurations:

    \begin{equation} 
       r_c^3 = \begin{cases} 
            -k_{c3}, & \text{if } \boldsymbol{\lambda} \le k_{c4} \\
            0 , & \text{otherwise}
        \end{cases}
    \end{equation}
where $k_{c3} \in \mathbb{R}^+$ is a scaling factor, $k_{c4} \in \mathbb{R}^+$ is the manipulability threshold. $r_c^3$ punishes the agent for getting close to singularity configurations and encourages the manipulator to stay dexterous to be able to perform the required tasks for safe debris removal.

The obstacle avoidance reward function $r_o$ is made up of two components, i.e., $r_o = r_o^1 + r_o^2$.

    i) A dense punishment for the distance between the points evenly distributed along the manipulator links $L_{ij}$ and the points on the surface of the rectangular base that is rigidly attached to the capture point $T_k$:

    \begin{equation}
        r_o^1 = -k_{o1} \max_{i,j,k}(e^-{\frac{(||L_{ij} - T_k|| - k_{o2})^2}{k_{o3}}})
    \end{equation}
where $k_{o1},k_{o2},k_{o3} \in \mathbb{R^+}$ are the scaling factors. $\max_{i,j,k}(\bullet)$ denotes iterating through the link points $L_{ij}$ and the surface points $T_k$ and entering the values into the function $\bullet$ and returning the largest output. When the manipulator is within a safe distance from the rectangular base, $r_o^1$ is negligible. But when the manipulator enters a danger region with the rectangular base, $r_o^1$ rapidly increases and the agent receives a dense punishment to discourage collision.

    ii) A sparse punishment for collision between the manipulator and the rectangular base:

    \begin{equation}
        r_o^2 = \begin{cases} 
            -k_{o4}, & \text{if } ||L_{ij} - T_k|| \le k_{o5} \\
            0 , & \text{otherwise}
        \end{cases}
    \end{equation}
where $k_{o4} \in \mathbb{R^+}$ is a scaling factor and $k_{o5} \in \mathbb{R}^+$ is a distance threshold. The activation of this sparse punishment represents an unwanted collision between the space manipulator and the rectangular base, which triggers the termination of the episode.

\subsection{Multi Critic Network}
This paper proposes the use of two critic networks in order to handle the multi-requirements of safe debris removal. In line with the TD3 agent, there are two capture critic networks that evaluate the agents tracking. The capture critics approximate the state action function $Q_{ck}:S_c \times A \xrightarrow{} \mathbb{R}$ with the target counterparts $Q'_{ck}:S_c \times A \xrightarrow{} \mathbb{R}$ for $k = 1,2$ that are parameterized by $\phi_{ck},\phi'_{ck}$ respectively. $\phi_{ck}$ are updated by minimizing the loss function $L_{ck}$  for $k = 1,2$

\begin{equation}
    L_{ck} = \frac{1}{N}\sum_{i=1}^{N}({y_{ci} - Q_{ck}(s_{ci},a_i|\phi_{ck})})^2
\end{equation}

\begin{equation}
    y_{ci} = r_{ci}(t) + \gamma \min_k(Q'_{ck}(s_{ci}(t+1),clip(\mu'(s_i(t+1)|\theta') + \epsilon_2)|\phi'_{ck}))
\end{equation}

There are two obstacle critics that focus on collision avoidance that approximate the state action function $Q_{ok}:S_o \times A \xrightarrow{} \mathbb{R}$ with the target counterparts $Q'_{ok}:S_o \times A \xrightarrow{} \mathbb{R}$ for $k = 1,2$ that are parameterized by $\phi_{ok},\phi'_{ok}$ respectively. $\phi_{ok}$ is updated by minimizing the loss function $L_{ok}$ defined by (39) and (40) for obstacle states $s_o \in S_o$ and obstacle reward $r_o$.
$\phi'_{ck},\phi'_{ok}$ are updated based on (27). Although each critic only sees the relevant portion of the total state $(s_c,s_o)$ that is required to compute the respective Q value $(Q_c,Q_o)$, both critics are fed the complete action outputted by the actor based on the current full state $s \in S$. In order to weight the importance of capture and obstacle avoidance, which are tasks that often contrast with each other, we define the constants $\alpha_c,\alpha_o >1$ such that $\alpha_c + \alpha_o = 1$. The actor weights $\theta$ are updated by performing gradient ascent on the following function

\begin{equation}
    J = \frac{1}{N}\sum_{i=1}^N\alpha_c \min_k(Q_{ck}(s_{ci},a|\phi_{ck})) + \alpha_o \min_k(Q_{ok}(s_{oi},a|\phi_{ok}))
\end{equation}

\subsection{Priority Experience Replay}

Standard TD3 samples experiences uniformly from a replay buffer $\mathcal{B}$, which prevents overfitting but can slow convergence. In tasks such as safe debris removal, some experiences (e.g., end-effector near target) are more informative. We propose a {Priority Experience Replay (PER)} framework that ranks episodes by success and cumulative reward $R_t$, storing experiences from the top 50\% in a priority buffer $\mathcal{B}_p$.

During training, minibatches of size $N$ are formed by sampling $\lambda N$ from $\mathcal{B}_p$ and $(1-\lambda)N$ from $\mathcal{B}$. The sampling rate $\lambda$ is adaptive based on the recent success rate $\psi$: high $\lambda$ when $\psi$ is low to exploit valuable experiences, gradually decreasing as $\psi$ increases to ensure robust learning. The update rule is

\begin{equation}
    \lambda = 
    \begin{cases} 
        \lambda_{max}, & \psi < 0.4, \\
        2.5\lambda_{max}(0.8 - \psi), & 0.4 \le \psi \le 0.8, \\
        0, & \psi > 0.8. 
    \end{cases}
\end{equation}

Episodes are grouped into intervals of $E_p$ episodes. After each interval, episodes are ranked by success and total reward, and experiences from the top 50\% are transferred to $\mathcal{B}_p$. Algorithm~1 summarizes the PER procedure.

\begin{algorithm}
\caption{PER Sampling Algorithm}
\begin{algorithmic}[1]
\STATE Initialize $\mathcal{B}, \mathcal{B}_p, \lambda, \psi$, episode info $\delta$
\FOR{$e = 1$ to $E$}
    \STATE Execute actions, store experiences in $\mathcal{B}$
    \STATE Sample $\lambda N$ from $\mathcal{B}_p$ and $(1-\lambda)N$ from $\mathcal{B}$ for minibatch
    \STATE Update TD3 agent
    \IF{$e \mod E_p = 0$}
        \STATE Rank latest $E_p$ episodes by success and $R_t$
        \STATE Update $\psi$ and $\lambda$
        \STATE Transfer top 50\% experiences to $\mathcal{B}_p$
        \STATE Reset $\delta$
    \ENDIF
\ENDFOR
\end{algorithmic}
\end{algorithm}

\section{Experiments And Results}
This section provides the simulation results. The simulation was performed in Simulink using Simscape to model multi-body systems. The manipulator used is a 7 DOF Redundant Kuka LBR iiwa attached to a cubic base forming the space manipulator. The agent was trained in two tasks. In task 1, the obstacle was absent and the agent was trained to follow the randomly moving capture point. In task 2, the capture point is rigidly attached to a rectangular base. The agent is trained to acquire the capture point which is now stationary but now with the rectangular base that is rotated to partially obscure the path between the capture point and the end effector. The manipulator is always initialized in a dexterous configuration, and the capture point is always initialized within the manipulator workspace. Figure 1 shows a block body diagram that conveys the general process of the simulation.

\begin{figure}[h] 
    \centering 
    \includegraphics[width=0.5\textwidth]{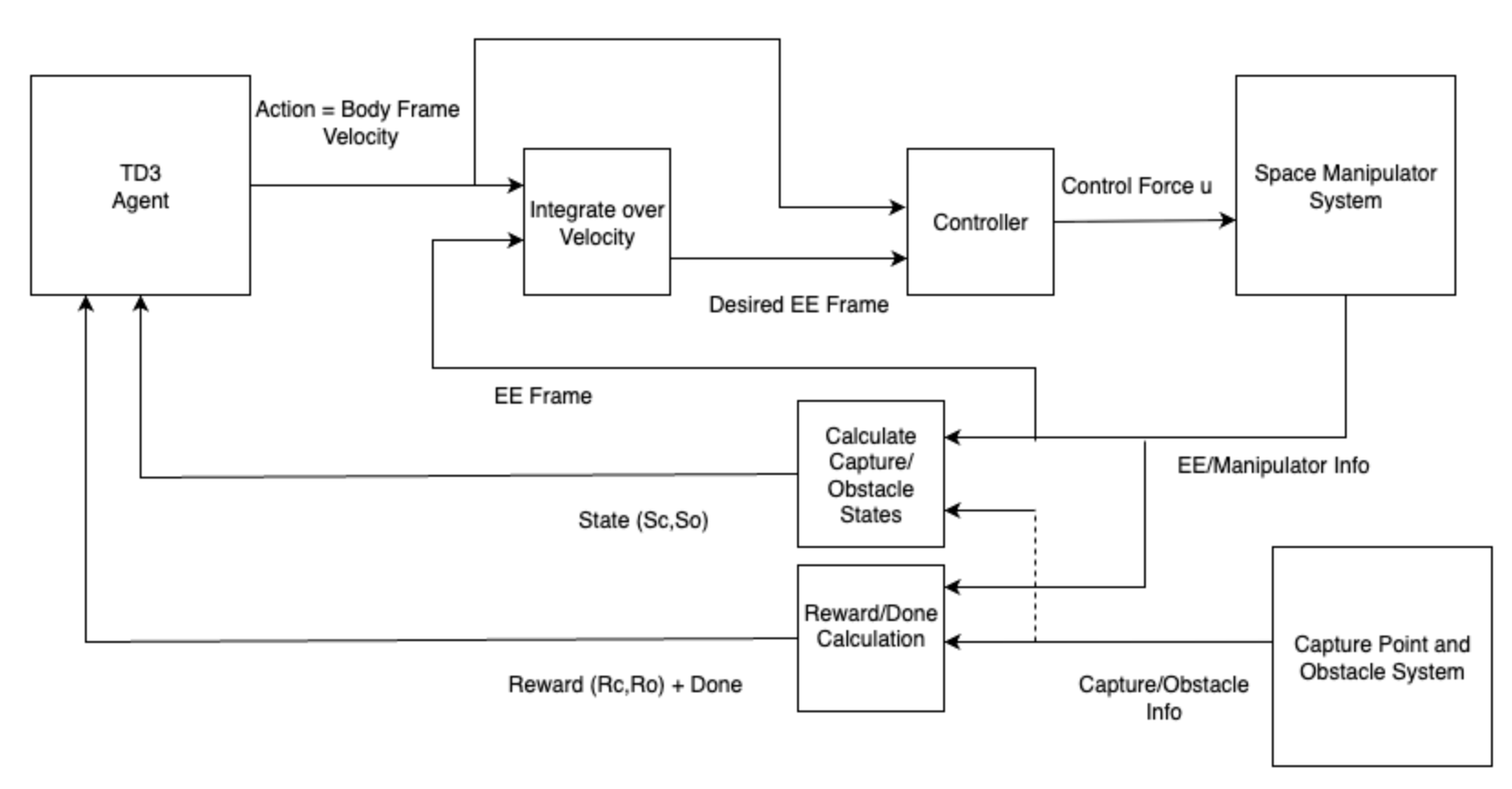}
    \caption{Simulation Block Diagram}
\end{figure}

\begin{figure}[h] 
    \centering 
    \includegraphics[width=0.25\textwidth]{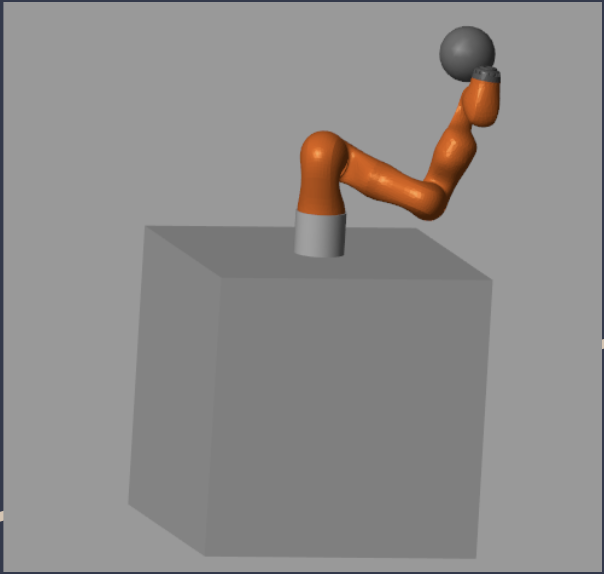}
    \caption{Task 1 Simulation Environment}
\end{figure}

\begin{figure}[h] 
    \centering 
    \includegraphics[width=0.25\textwidth]{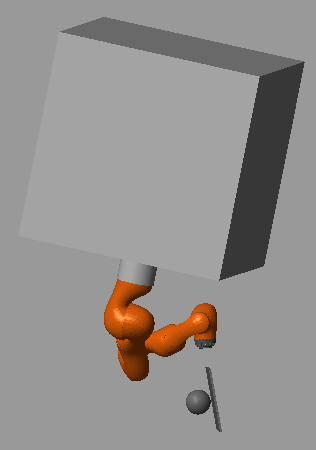}
    \caption{Task 2 Simulation Environment}
\end{figure}

\subsection{Task 1}

In task 1 tracking was the main objective, and therefore only the capture critics $Q_{ck}$ are trained, and the actor network $\mu(s_c)$ only sees the capture state $s_c$. The rectangular base acting as an obstacle between the end effector and the capture point is absent, allowing the TD3 agent to focus solely on tracking. Figure 2 shows the environment of task 1. The capture point is translating randomly in 3D space at a speed of $0.02 m/s$. In task 1, successful contact between the end effector and capture point does not terminate the episode, but rather the agent is rewarded for staying in contact with the capture point, allowing the agent to gain more experience tuple tracking. A baseline TD3 agent was used in task 1 without any PER. The actor network consisted of two hidden layers each with 800 neurons. The two capture critics had two hidden layers each with 400 neurons.

Task 1 was performed 4 times, Figure 4 shows the capture reward graph of all the runs with the black curve representing the average between the four runs. As seen in figure 4, on average the TD3 agent remains unchanged for the first 500 episodes, but then drastically improves. The TD3 agent converges right after 1200 episodes, and once the TD3 agent converges, it remains consistent for the remaining training episodes.

\begin{figure}[h] 
    \centering 
    \includegraphics[width=0.45\textwidth]{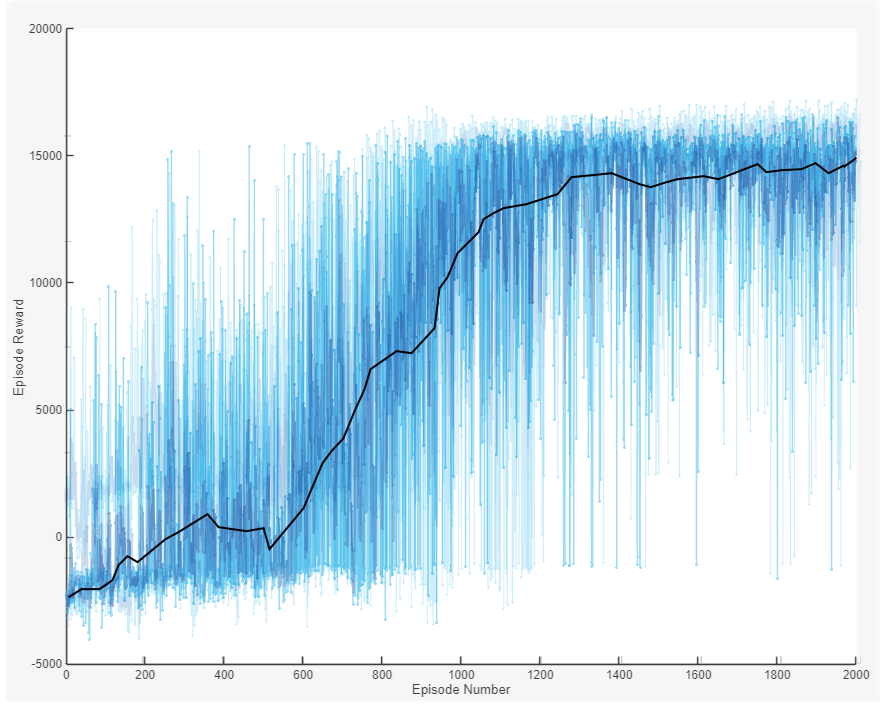}
    \caption{Task 1 Capture Reward Graph}
\end{figure}

\subsection{Task 2}

In task 2 the capture point is rigidly attached to a rectangular base that acts as the obstacle impeding the path between the end effector and the capture point. In this task, the capture point is completely stationary. Obstacle critics $Q_{ok}$ are introduced and trained. Notice that with the addition of $Q_{ok}$, the states are no longer just $s_c$, but rather the concatenation of $s_c$ and $s_o$. Hence, a new actor network $\mu(s_c,s_o)$ needs to be initialized. Furthermore, while $Q_{ok}$ is learning, previously trained capture critics $Q_{ck}$ are frozen and are not being trained, $Q_{ck}$ are only used to train the new actor network. In this task, contact between the end effector and the capture point will trigger the termination of the episode. Task 2 is performed with the proposed multi-critic TD3 agent with PER. The architecture of the actor and capture critics remained the same as in task 1, the dual obstacle critics introduced had two hidden layers each with 600 neurons. Figure 3 shows the simulation of task 2.

Task 2 was run 4 times and figure 5 shows capture reward of all the runs with the black curve representing the average of all the curves. Figure 6 displays the percentage of successful episodes for every $Ep$ episodes during the learning process. Based on figures 4 and 5, the TD3 agent quickly learns to effectively move around rectangular base acting as the obstacle to successfully acquire the capture point. It only takes the TD3 agent around 200-300 episodes to converge and once again the TD3 agent stays consistent throughout the remainder of the training process. The speed of convergence is hypothesized due to the multi critic and PER frameworks which proved to be effective in \cite{Timothy}.

\begin{figure}[h] 
    \centering 
    \includegraphics[width=0.45\textwidth]{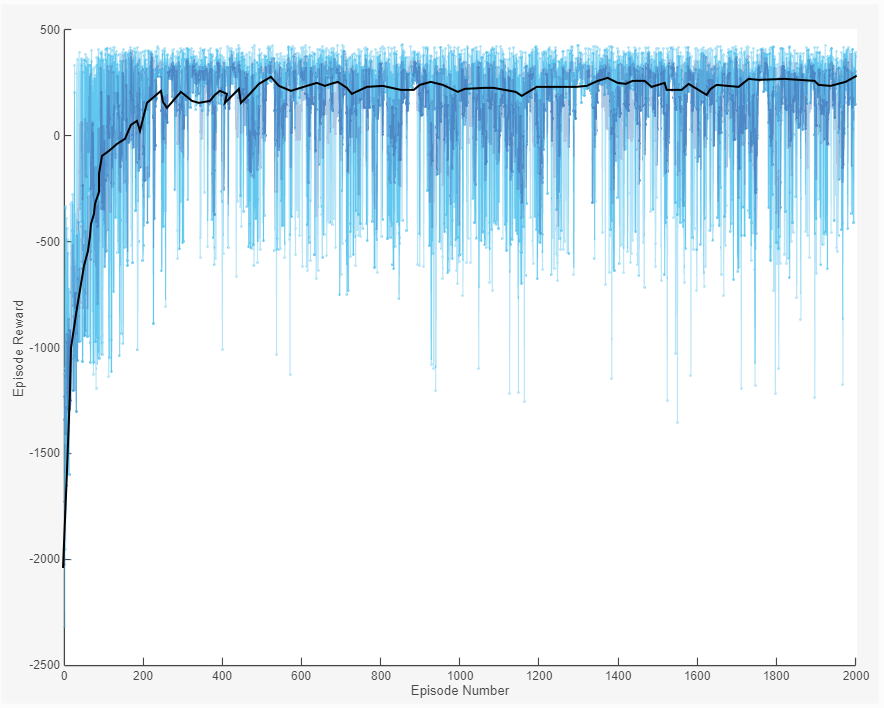}
    \caption{Task 2 Capture Reward Graph}
\end{figure}

\begin{figure}[h] 
    \centering 
    \includegraphics[width=0.5\textwidth]{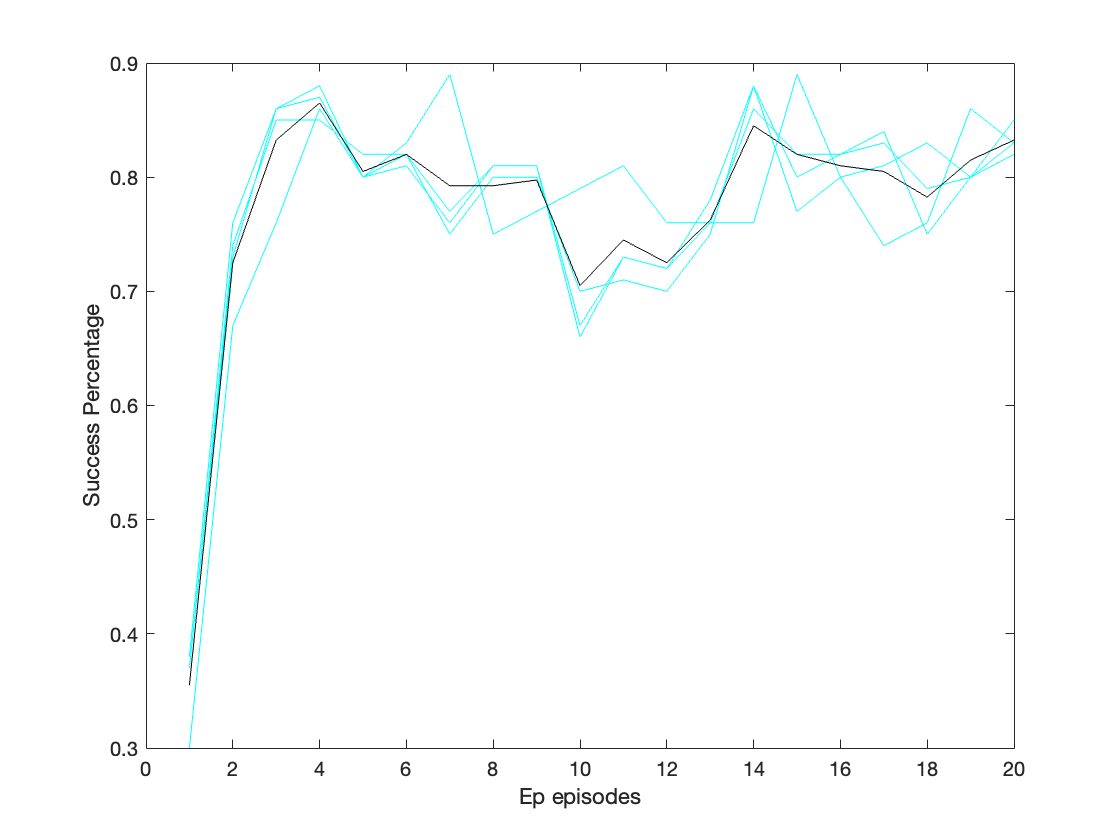}
    \caption{Task 2 Success Percentage With PER}
\end{figure}

\section{Conclusion}
This paper was a culmination of previous works and served to combine the DRL training methods in \cite{Timothy} with the robust controller developed in \cite{Borna_Control2} and \cite{Borna_Control}. In order to accomplish the task of safe debris removal, we utilize a RL TD3 agent to act as the obstacle free trajectory planner for the end effector and proposed the usage of multi-critic networks and PER to train the agent. The combination of the controller, multi critic networks, and PER proved to be effective and capable in 3D spatial environments with obstacle avoidance considerations which are more complex and realistic when compared to 2D planar space manipulators that only focus on the tracking, which are more common in the current literature. In the future, we propose conducting the experiment with a real robotic manipulator. 

\printbibliography 

\end{document}